# Whole-Slide Image Focus Quality: Automatic Assessment and Impact on AI Cancer Detection

**Authors and emails**
Timo Kohlberger (tkohlber@google.com)[1]
Yun Liu[1]
Melissa Moran[1]
Po-Hsuan (Cameron) Chen[1]
Trissia Brown[1]
Craig H. Mermel[1]
Jason D. Hipp[1]
Martin C. Stumpe (mstumpe@google.com)[1]

**Affiliations**
[1]Google AI Healthcare, 1600 Amphitheatre Parkway, Mountain View, CA, USA

**Corresponding author:**
Timo Kohlberger (tkohlber@google.com)

# Abstract

**Background:** Digital pathology enables remote access or consults and powerful image analysis algorithms. However, the slide digitization process can create artifacts such as out-of-focus (OOF). OOF is often only detected upon careful review, potentially causing rescanning and workflow delays. Although scan-time operator screening for whole-slide OOF is feasible, manual screening for OOF affecting only parts of a slide is impractical.

**Methods:** We developed a convolutional neural network (ConvFocus) to exhaustively localize and quantify the severity of OOF regions on digitized slides. ConvFocus was developed using our refined semi-synthetic OOF data generation process, and evaluated using real whole-slide images spanning 3 different tissue types and 3 different stain types that were digitized by two different scanners. ConvFocus's predictions were compared with pathologist-annotated focus quality grades across 514 distinct regions representing 37,700 35x35µm image patches, and 21 digitized "z-stack" whole-slide images that contain known OOF patterns.

**Results:** When compared to pathologist-graded focus quality, ConvFocus achieved Spearman rank coefficients of 0.81 and 0.94 on two scanners, and reproduced the expected OOF patterns from z-stack scanning. We also evaluated the impact of OOF on the accuracy of a state-of-the-art metastatic breast cancer detector and saw a consistent decrease in performance with increasing OOF.

**Conclusions:** Comprehensive whole-slide OOF categorization could enable rescans prior to pathologist review, potentially reducing the impact of digitization focus issues on the clinical workflow. We show that the algorithm trained on our semi-synthetic OOF data generalizes well to real OOF regions across tissue types, stains, and scanners. Finally, quantitative OOF maps can flag regions that might otherwise be misclassified by image analysis algorithms, preventing OOF-induced errors.

# Background

Digital pathology is advancing into clinical workflows [1], enabled by the recent regulatory approval of the first Whole-Slide Image (WSI) scanner for primary diagnosis in the U.S. [2], as well as wider availability of cheaper storage and large technical infrastructure to manage gigapixel-sized image files. Digitization has several compelling use cases: archiving, telepathology for remote consults or diagnosis, teaching, and increasingly, facilitation of powerful image analysis algorithms [3,4]. The process of digitization, however, can add artifacts to the imaging process, including color or contrast variations and out-of-focus (OOF) areas. These artifacts, particularly OOF areas, can hinder the rendering of accurate diagnoses by pathologists [1], or impact the accuracy of automated image analysis [5].

We here distinguish between three general categories of OOF in decreasing severity: *global OOF* that affects the entire slide, *regional OOF* that affects larger tissue patches, and *local OOF* that affects individual cells or subcellular structures (Figure 1). Global OOF can be caused by the whole-slide scanner erroneously focusing on coverslip debris. Regional OOF can be particularly prominent in specific tissue types (such as fat), or affect an entire section of the slide (e.g. "scan lane"). Lastly, local OOF is common because typical tissue sections of approximately 5 microns are thicker than the depth of field of standard digital pathology scanners at high magnification. While a potential solution is the use of multiple focal depths ("z-stacking"), such techniques currently increase scan time and file sizes to an impractical extent for routine use.

The different OOF categories have different impacts on the clinical workflow. Pathologists or histotechnicians may flag images with global OOF as low quality and order a rescan, which potentially results in reporting and workflow delays. Regional and local OOF can be much more difficult to detect consistently, particularly when reviewing a slide at low magnification. Rescans may only be requested after the pathologist has spent significant time reviewing other areas in the slide, before proceeding to evaluate specific areas of concern at higher magnification. Importantly, a technician can pre-screen all digital slides for global OOF, but that would be impractical for manual review for smaller OOF regions. OOF image artifacts can have even more severe consequences in automated image analysis by directly impacting detection and classification. Some studies found that systematic errors can be attributed to suboptimal focus quality, such as OOF germinal centers being mistaken for tumor metastases by an algorithm [5].

While every WSI scanner has built-in focus evaluation that can be used for automatic rescans of the affected regions or for quality reporting, there are several shortcomings in existing methods: (1) despite built-in auto-focus control, most currently available WSI scanners still produce scans with focus issues [6], (2) focus evaluation methods across scanners are different, inhibiting comparison across devices, (3) focus metrics are not typically exported to the user with detail regarding the spatial distribution of focus quality, and (4) evaluation does not take the clinical relevance of the focus quality into account. For example, cytology diagnoses that more strongly rely on subcellular details require a higher focus quality than diagnoses that are based primarily on tissue architectural patterns, such as prostatic adenocarcinoma Gleason grading.

A sizable body of work exists for automated focus quality assessment in microscopy images using manually engineered image features, e.g. [7]. Newer approaches have used deep convolutional neural networks [8], which automatically learn the discriminative features and yield higher accuracies [9][10]. The main challenge, particularly for the neural networks approaches, is the availability of training data to help generalization across a large variety of tissue morphologies and stain properties. To help generalization, previous works have generated simulated OOF examples using real in-focus images via synthetic Gaussian blurring [9], and additional perturbations such as brightness perturbations and artificial sensor noise [8].

This work further improves the synthetic data generation approach by more closely mimicking the image acquisition process of real OOF artifacts. These improvements allowed us to build a robust, highly sensitive and fine-grained OOF predictor (ConvFocus) that distinguishes a large spectrum of focus degrees across different tissue and stain types. Our approach provides a generally applicable metric that is highly concordant with manually annotated focus quality and provides information about focus quality across every region in a WSI. Moreover, we quantify the sensitivity of a leading tumor detection algorithm to focus quality, which has not been shown in prior related work.

# Materials and Methods

## Overview

Given a gigapixel-sized image of a whole pathology slide, our goal was to automatically detect and grade OOF regions at an accuracy level matching that of a pathologist across tissue, biopsy and stain types. To that end, we employed a convolutional neural network approach, which has shown superior performance over more classical machine learning approaches using hand-crafted features [9]. We term our approach ConvFocus (Figures 2-3).

## Training Data

Developing accurate neural networks requires large volumes of training data. Unfortunately, accurate delineation and grading of OOF areas in WSIs is a highly time-intensive task. For example, our pathologist spent 8-16 hours per slide to find, delineate and grade OOF regions (see the "Test Data" section for more details). Consequently, generating a sufficiently large pool of OOF training examples purely by manual annotation is impractical. Instead, we first manually assessed whether patches were completely in-focus or not. For each patch that was labelled "in-focus" by three independent raters, we then synthetically generated multiple additional OOF patches with varying but known degrees of OOF. The real in-focus image patches and the synthetically "de-focused" versions served as training examples. We next describe this process in detail.

### Whole-Slide Scans Used for Training

26,526 different slides spanning a large variety of tissue, biopsy and stain types were used to train and validate our convolutional neural network. Of these, 8,135 slides had surgical sectioning site information available in a structured form (Table 1). 26,099 slides were scanned with an Aperio AT2 (Leica Biosystems, Germany) at 40X magnification (pixel size: 0.251×0.251µm), using a semi-automatic scan mode, and 427 slides were scanned using a Hamamatsu NanoZoomer XR (Hamamatsu Photonics, Japan) also at 40X magnification (pixel size: 0.227×0.227µm).

## Manual Annotation of In-focus Image Patches

In-focus images were obtained by manual rating of random tissue-containing image patches from the training slides. First, tissue-containing regions were determined using a threshold. For every 128×128 pixel patch (34×34μm), it was labeled as tissue-containing if the average luma value $Y = 0.212 \cdot R + 0.715 \cdot G + 0.072 \cdot B$ fulfilled $0.0 < Y \leq 0.8$ (with R, G, B in the range [0,1]). Next, 300×300 pixel-sized patches were randomly sampled from tissue-containing areas. Of those the top 8 (for the AT2) and top 24 (for the NanoZoomer XR) patches with the lowest average luma value were picked for manual assessment of whether the patch was in-focus. More patches were selected from NanoZoomer XR images because fewer images from that scanner were available in our database, thus this was necessary to achieve a more balanced training set with respect to scanner. This procedure was designed to select the densest tissue parts, which in most cases provided the highest density of sharp image gradients and best exposed OOF issues.

These "densest" patches were then sent to trained non-pathologist raters for independent evaluation. Raters applied one of the following labels to each patch: "in-focus" if at least 75% of tissue area was in-focus, "out-focus" if at least 75% was OOF to any degree, and "undecided" otherwise. We selected approximately 166,000 patches that all three raters labeled as "in-focus" (i.e. consensus rating) to create semi-synthetic training data.

## Semi-synthetic Training Data Generation

With the rater-determined in-focus patches, we synthetically generated 30 different versions with increasing severity of OOF ranging from class 0 to class 29. Class 0 represented the original in-focus patches, classes 1-28 represented fine-grained increasing OOF magnitudes, and class 29 covered a large range of strong OOF artifacts.

Prior work used convolutions with 2-D Gaussian kernels with varying σ to simulate OOF. However, we hypothesized that real OOF artifacts in WSI more closely resemble the "bokeh" OOF effects in photography [11][12], which are generated by overlapping superpositions of the lens or the aperture shape. This motivated experimentation with a circular convolution kernel with varying radius r, which we will refer to as "Bokeh blurring".

To determine the relationship between the OOF class label and the blurring kernel size (i.e. Gaussian's σ or Bokeh's circular convolution stencil radius r), we visually assessed the synthetic blurring magnitudes for different mappings (Figure 4). Compared to a linear relationship, an exponential one was found to better represent the intermediate OOF grades. Specifically, for OOF classes $c \in [1, 28]$, σ for Gaussian blurring was chosen randomly from intervals $[\,0.926 \cdot exp(\,3 \cdot \frac{c-1}{28}),\ 0.926 \cdot exp(\,3 \cdot \frac{c}{28})\,)$, and the radius of the Bokeh blurring kernel from $[\,1.4 \cdot exp(\,3 \cdot \frac{c-1}{28}),\ 1.4 \cdot exp(\,3 \cdot \frac{c}{28})\,)$, respectively. For class $c = 29$, σ values were chosen from $[\,0.926 \cdot exp(\,3),\, 132\,)$, and radius from $[\,1.4 \cdot exp(\,3),\, 200\,)$, respectively. To help the network capture the full OOF interval range during training, varying blurring levels were randomly sampled from within the class-specific intervals, instead of only using the intervals' center values. The scaling factors 0.926 and 1.4, and maximum values 132 and 200 were found to yield similar blur strengths for either method and the same OOF class. These scaling factors were determined by minimizing the sum of squared differences between RGB values of the images over all OOF classes (Figure 4).

## Post-blurring JPEG and Noise Artifacts

In initial experiments, convolutional neural networks trained on Bokeh or Gaussian-blurred synthetic examples yielded poor prediction accuracy on real OOF images; erroneously predicting almost all OOF test patches as in-focus (see Results). We hypothesized that this was caused by the artificial smoothing removing other types of real artifacts. For example, grid-like artifacts at the edges of scan lanes and JPEG blocks are smoothed out in artificially blurred images, but would be present in real OOF images. Thus, several categories of other artifact types were re-added after synthetic blurring (Figure 5).

Visual comparison at high magnification of synthetically blurred real in-focus images to real OOF revealed other artifacts beyond OOF: pixel noise, likely from the image sensor; and Joint Photographic Experts Group (JPEG) compression artifacts, likely from the lossy JPEG compression applied by scanners post digitization (scanner quality settings in our data ranged from 70-90). In synthetically blurred images however, both of these artifacts ranged from faint to absent, depending on the synthetic blurring magnitude, even if they were present in the original in-focus input images. This is because both pixel noise and JPEG artifacts typically consist of high spatial frequencies, which are diminished by blurring.

Consequently, simulated JPEG compression artifacts were added back into the synthetically blurred images, implemented via JPEG encoding and decoding and an image quality parameter chosen between 70 and 90%. Pixel noise was added because image sensors collect Poisson-distributed noise. Therefore, pixel-wise Poisson noise was simulated by mapping each color channel value $x_c \in [0,1]$, $c \in \{R,G,B\}$ to a noisy version $x'_c$ via: $x'_c = P(x_c/s) * s$, with $P$ referring to the Poisson distribution, and $s$ inversely controlling the signal-to-noise ratio. As the latter was observed to vary significantly between different scanners and objective magnifications, the noise portion during training was varied by randomly sampling s from the interval [0.01, 64.0] for each training patch.

## Test Data

The OOF algorithms were evaluated using pathologist-graded real OOF artifacts (as opposed to synthetic OOF in the training set). Specifically, 3 prostate resection and 4 lymph node biopsy slides were used, including Hematoxylin and Eosin (H&E) and immunohistochemistry stains. Each WSI was scanned using two scanners at 40X magnification: a Leica Aperio AT2 (pixel size: 0.251×0.251µm) using its semi-automatic mode, and a Hamamatsu NanoZoomer S360 scanner (pixel size: 0.230×0.230µm) using its fully automatic mode. This resulted in two test sets: 7 WSIs per scanner.

A pathologist then manually and non-exhaustively identified, delineated, and graded in-focus and OOF regions, using integral grades ranging from 0 (in-focus) to 6 (very strong OOF), see Figure 6. Half-grades (e.g. 1.5) were occasionally used when the degree of OOF was interpreted as between integral grades. Annotations were corrected by the pathologist as desired (blinded to algorithm predictions), to help achieve grading consistency across the different tissue and stain types, and scanner models. The delineated regions were rasterized using a 128×128 pixel-spaced grid to enable direct comparison with the predictions of the OOF classifier (see below). To ensure label purity, only patches completely contained within an delineated OOF region were used. These annotations produced a total of 37,715 patches.

In addition to pathologist annotations, we collected "z-stack" scans to further evaluate ConvFocus performance on a set of images that have real OOF with a consistent pattern. We digitized a lymph node biopsy slide with z-axis increments of 0.4μm, and spanning +4μm to -4μm (relative to the scanner-determined in-focus depth), see Figure 10(a). This produced a total of 21 WSIs of the same glass slide.

# Convolutional Neural Network-Based Algorithm

## Network Architecture

Our convolutional neural network architecture was a series of convolutions and pools, equivalent to the Inception (v3) architecture [5,13] truncated at a lower layer ("MaxPool_3a_3x3" [13]) and with a reduced number of filters per layer ("depth_multiplier"=0.1) to reduce computation. For this task, we empirically observed that neither modification deteriorated performance (Results). The last layer was attached to a 30-way classification layer to predict the 30 OOF classes. To handle the large image sizes of each WSI, we adopted a patch-based approach. Small crops of 139×139 pixels-sized patches (corresponding to approximately 35×35 μm at 40X magnification) for each WSI were used for training. To help improve performance, we applied data augmentation techniques described in [5], randomly perturbing the orientation, brightness, contrast, hue, saturation, and adding random translational jitter. The network was trained using the softmax cross-entropy loss function, using the same learning rate schedule and other hyperparameters described in [5].

## Inferring OOF Heatmaps on WSIs

To obtain algorithm predictions of OOF for every region on the slide, we applied ConvFocus in a sliding window fashion across each WSI to produce algorithm predictions for every 128×128 patch at 40X magnification. This stride was chosen to match the stride used for cancer detection (see below), and is adjustable. The final predicted OOF class for each patch was the class with maximal probability ("activation") in the final layer of the network. To visualize these predictions, we used the "jet" colorspace, ranging from blue for class 0 (in-focus) to red for class 29 (strongly OOF) (Figure 3).

## ConvFocus Evaluation

Each algorithm's performance was evaluated by comparing the predicted OOF class (in the range 0 to 29) to the corresponding pathologist-annotated OOF grades (0 to 6 in half-point increments, 13 grades overall) among all patches with annotations within the two test sets (one for each scanner: AT2 and S360). However, the numbers of patches across the 13 different pathologist-assigned grades were unevenly distributed within each test set, and the distributions differed between the two test sets. Therefore for each test set and grade, 3,000 patches were randomly sampled with replacement to obtain evenly distributed classes.

Similar to prior work [9], the Spearman's rank correlation coefficient was used as the main evaluation metric. To assess the statistical significance of the correlation, we used a two-sided test [14]. Next, we computed a linear regression to evaluate deviation of the intercept from 0.0, since an "in-focus" patch (annotated grade: 0.0) should ideally be predicted as "in-focus" (ConvFocus class: 0). The regression's slope was used to assess deviations from 4.83 (=29/6), which reflects a linear mapping of the ConvFocus OOF class range of [0,29] to the graded OOF range of [0,6].

In addition, ConvFocus was further evaluated on the z-stack test set by plotting the predicted OOF class against the z-level for a number of patches. Two aspects were qualitatively assessed. First, we checked for the presence of a "v"-shaped plot indicating that predicted OOF class increases (towards poorer focus) as the z-level goes further from in-focus (in either direction). Second, we checked if the "in-focus" plane (z=0) was generally predicted as the lowest OOF class.

## Measuring Cancer Detection Algorithm Performance As a Function of Focus Quality

After validating ConvFocus, we used the algorithm to study the impact of OOF (both real and synthetic) on the performance of a cancer detection algorithm on the publicly available Camelyon 2016 challenge test dataset [15], consisting of 80 non-tumor and 48 tumor slides with pixel-level annotations of tumor locations. In the first experiment (real OOF), we employed the current best-performing breast cancer metastasis detector "LYNA" [5,16], which we configured to provide binary predictions at the same granularity as the OOF classifier. That is, for each image patch, predictions from both the breast metastasis and ConvFocus were available. Next, we stratified the patches by predicted OOF class. We then measured the LYNA algorithm's performance for patches in each OOF class, merging the higher OOF classes to ensure sufficient numbers of both tumor and non-tumor patches. In a second experiment, we added artificial Bokeh blur at a selected strength to all patches in the test set, ran LYNA on all patches, and evaluated LYNA's performance. We repeated this for a range of Bokeh-blur strengths. In both experiments, we used the patch-level area under receiver operating characteristic curve (AUC [17]). Confidence intervals for the patch-level AUC were computed using the non-parametric bootstrap approach [18], with n=500 samples at the slide level.

# Results

## ConvFocus Evaluation

Sample visualizations of ConvFocus's predictions are shown in Figures 3 and 7, depicting regional OOF patterns such as OOF scan lanes and OOF tissue edges. Quantitative evaluation using a total of 37,715 pathologist-graded OOF patches is reported in Table 2 and visualized in Figure 8, showing high Spearman rank correlations above 0.8 (p<0.001) between pathologist-graded OOF and ConvFocus predictions for slides scanned on both scanners tested: AT2 and S360.

Each step in our proposed semi-synthetic data generation approach was derived using qualitative assessments of whether regions with strong OOF were appropriately detected (Figure 7). To quantify the improvements derived from each step, we also performed ablation experiments using several configurations (Table 2). Using Bokeh synthetic blurring instead of Gaussian yields only a small improvement in correlation metrics, but a large increase in the regression's slope parameter values. Since the synthetic Gaussian and Bokeh blurring magnitudes were visually and quantitatively aligned (Methods), similar slope values were expected for the Gaussian-trained configuration. Instead, the Gaussian configuration classified patches annotated with OOF grades 5.0-6.0 as mid-range OOF classes 14-20 (Figure A1) compared to 20 and above with the Bokeh configuration (Figure 8). Moreover, the majority of patches with no or weak OOF (annotated grades 0.0-1.5) were predicted as OOF classes 0-4 by the Bokeh configuration (ConvFocus), compared to 5 and higher with the Gaussian configuration.

We next assessed ConvFocus's predictions for patches from a set of 21 "z-stack" images (Figure 9). Almost all patches were predicted to be in-focus at a z-level, though not necessarily at z=0 (Figure 10). A strong "v"-shaped trend was also observed: patches were generally predicted to be more OOF at z-levels further from the predicted in-focus z-level (on average at z=0). However, using the Gaussian configuration, few patches were predicted to have OOF class < 5 (Figures A2a and A3). Similarly, the configuration with Bokeh blur but linearly increasing convolution mask size (instead of exponential) resulted in less fine-grained sensitivity towards weak OOF degrees (Figures A2b and A4).

## Impact of Focus Quality on Cancer Detection Algorithm Performance

Figure 11 shows the effect of OOF on patch-level AUC for the LYNA metastatic breast cancer detector. Results for the first experiment (effects of real OOF) are shown in black dots with gray bars that indicate 95% confidence intervals. Since few image patches in the test set were predicted as higher OOF classes, we bucketed the classes into 5 groups: 0-4, 5-9, 10-14, 15-19, and 20-29. Results indicated a gradual drop in cancer detector performance with predicted OOF, as evidenced by decreasing AUC values and the associated upper bounds of the confidence intervals.

To assess the effects of finer-grained OOF, we conducted a second experiment: synthetically adding OOF, as depicted by the blue curve. In contrast to the first experiment, since synthetic blur at each specific degree was applied uniformly across all slides, no class-merging was necessary. Results similarly showed the gradual and consistent drop in cancer detection with increasing OOF.

# Discussion and Conclusions

We have developed and evaluated an automated focus quality detector (ConvFocus) to accurately locate and classify OOF regions as small as 32×32μm in gigapixel-sized WSIs. Our algorithm correlates well with pathologist-annotated focus quality grades on real OOF regions, and across multiple scanners despite different imaging characteristics (e.g. pixel size in the S360 is 9% smaller than in the AT2). Because different scanners do not generally digitize the same slide locations with the same focus quality, the actual image patches used in this evaluation are different. As a result, a direct and fair inter-scanner performance comparison is not possible.

In a second evaluation using z-stacks to produce digitized images of varying OOF degrees, ConvFocus similarly produced sensible trends of increasing predicted OOF classes in both directions away from the scanner-determined in-focus plane. In this evaluation, while the scanner-determined in-focus plane (z=0) was on average predicted to be close to or the most in-focus, this was not necessarily true of all image patches. Further analysis is needed to determine whether this is caused by inaccuracies in focus detection by the scanner's algorithm, or by ConvFocus. The asymmetry in this "v"-shaped trend may be because the z-levels' span of 8μm exceeds the thickness of a tissue section (5μm), and thus some z-levels will overlap the coverslip or glass slide. This effect may be further exacerbated if the z=0 plane is not centered within the tissue section.

A significant advantage of our semi-synthetic data generation approach is an easy extension to additional scanner models, tissues, and stain types. Specifically, input data can be generated by non-expert raters, improving the speed and cost of data collection. Our results indicate that our method of synthesizing OOF examples of different degrees enables better generalization to real OOF images. In terms of granularity, ConvFocus was developed for fine-grained discrimination between 30 OOF classes, significantly more than both our pathologist's grading scale and prior works: 6 classes [9] and 12 classes [8,9]. When the granularity was further increased in our experiments (e.g. to 40 classes, data not shown), training accuracy decreased, indicating that 30 classes is close to a meaningful upper bound of granularity. Our qualitative (Figure 4) and quantitative (Figure A1) analyses also found that human perception of OOF followed an exponential relationship with blur magnitude, as opposed to linear. The exponential relationship expands the granularity of fine-grained OOF classes, and is particularly relevant for use cases like lymphoma diagnosis, where even mild OOF may interfere with the ability to resolve crucial nuclear details.

Prior works in synthetic data generation for training focus quality predictors have leveraged Gaussian blurring for histopathology [9] and Gaussian blurring with Poisson noise for cytology specimens from transmission light microscopes [8,9]. By contrast, our work systematically demonstrates the value of adding Poisson noise and JPEG compression artifacts in histopathology, and using a synthetic blur (Bokeh) that more closely mimics real OOF. Our results indicate that generating synthetic OOF data using the Gaussian blur underestimates strong OOF. Based on these improvements, we showed that focus quality can be accurately predicted on a fine-grained scale using fairly shallow neural networks, which improves the speed of applying ConvFocus to large WSIs.

We further investigated the impact of focus quality on an otherwise accurate algorithm's interpretation [5], showing that algorithm performance was lower in image patches identified by ConvFocus as OOF. Furthermore, our controlled experiment of synthetically "de-focusing" the image patches showed similar trends, demonstrating that synthetic OOF causes degradations in algorithm performance. These findings suggest that focus quality can negatively impact algorithms, and that focus quality should be comprehensively assessed at the region level to avoid both false positive and false negative interpretations by an algorithm.

The present work contains some limitations. Firstly, the cost of manual grading of focus quality has limited our test dataset size and diversity in terms of stains, tissue types, scanner models, as well as number of annotators. Further work will be needed to further validate the generalizability of ConvFocus against multiple rater annotations [9], and quantifications of z-stack prediction performance. Second, using our algorithm to compare focus quality across different scanners was confounded by the fact that different image patches are identified by our pathologist as OOF. Additional work will be needed to obtain scanner-agnostic or tissue-independent metrics of focus. Lastly, we have focused on the high magnification, 40X, which most clearly shows OOF artifacts. The extension to the next-lower 20X magnification is the subject of future work.

# Financial support and sponsorship

This study was performed and funded by Google Research.

# Conflicts of interest

T.K., Y.L., M.M., P.C., J.H., C.M., and M.S. are employees of Google, LLC and own Alphabet stock. T.B. is compensated by Google, LLC for the participation as pathologist in these studies.

# Acknowledgments

We would like to thank members of the Google AI Pathology team for software infrastructural and logistical support, and slide digitization. Gratitude also goes to Daniel Fenner for assistance with Bokeh blur, and Samuel J. Yang for helpful discussions.

# Figures

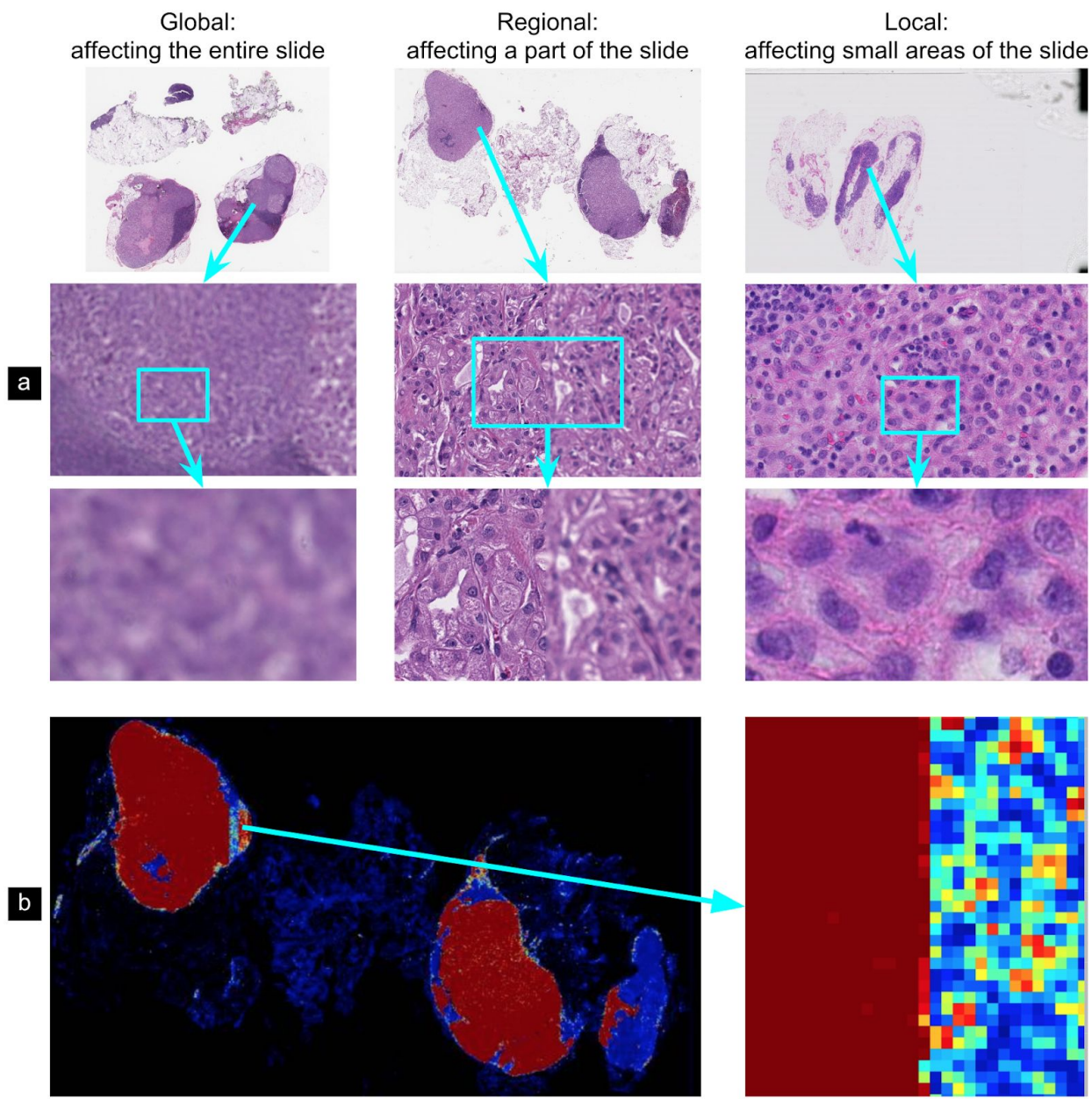

**Figure 1: Examples of different types of out-of-focus (OOF).** (a) Examples of different types of OOF: "global" affecting the whole slide, "regional" affecting a large expanse on a slide, and "local" affecting small tissue or cellular areas. (b) Example of the OOF scan lane in the regional OOF above (middle), that causes a striking artifact in a leading cancer detection algorithm. Left: the heatmap visualization of cancer predictions; black indicates non-tissue-containing regions; and other colors range from blue (non-tumor) to red (tumor). The OOF scan lane is predicted to be non-tumor. Right: a zoom in of the OOF scan lane boundary.

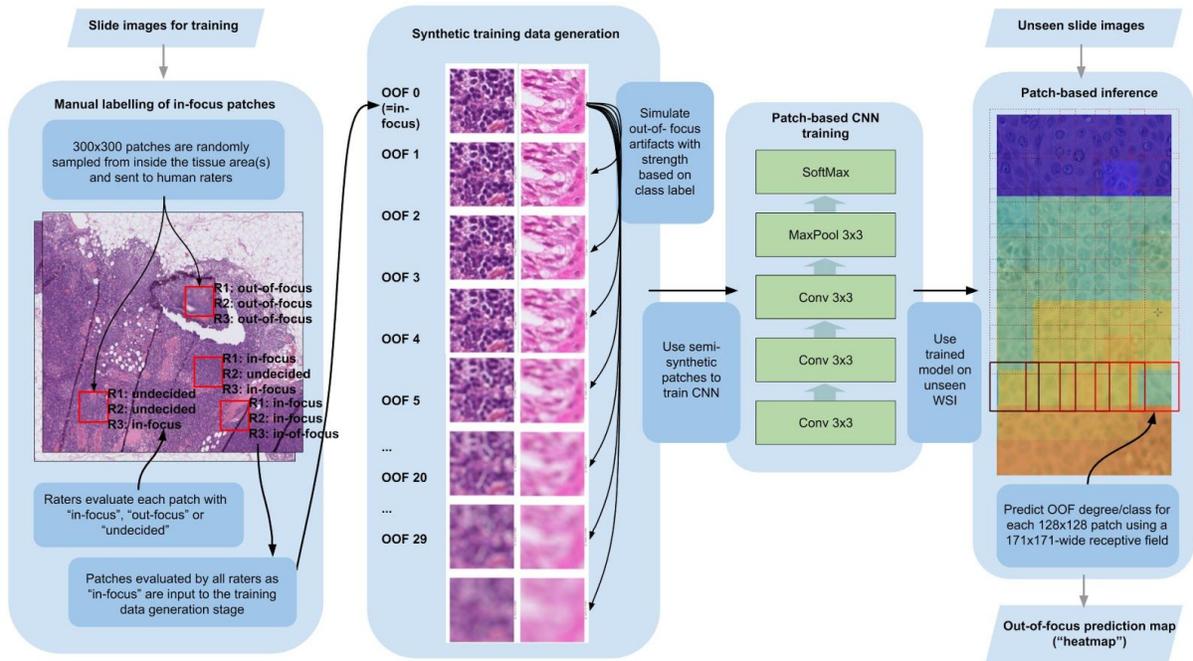

**Figure 2:** Overview of our convolutional neural network (CNN) approach to automated out-of-focus (OOF) grading: ConvFocus.

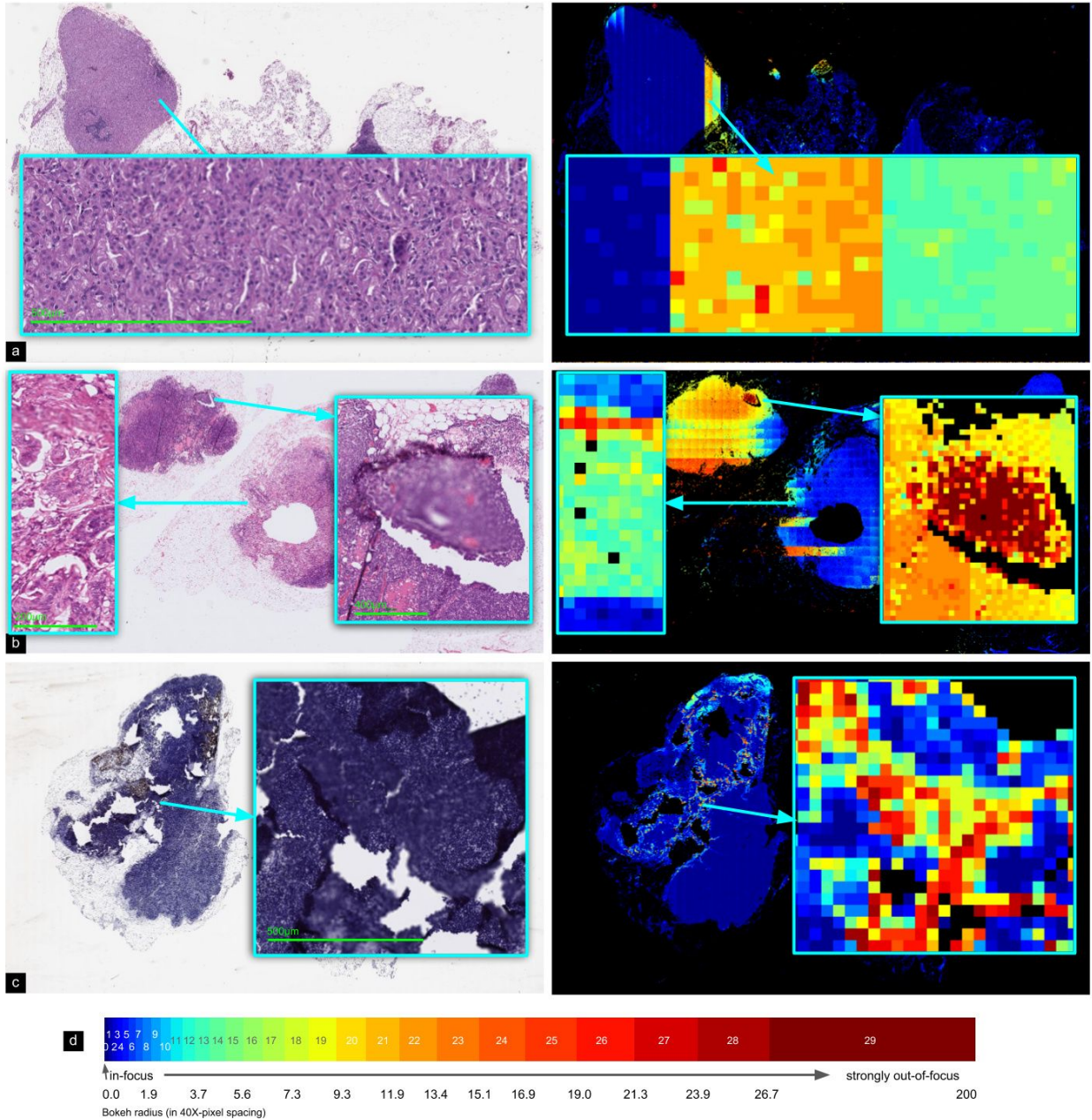

**Figure 3: Sample predictions from ConvFocus.** Left: Whole-slide image (WSI) of a lymph node biopsy from a colon cancer case, exhibiting both regional (along scan lanes) and local OOF at varying degrees. Right: predicted OOF classes expressed using a "jet" colormap from blue (in-focus) to red (strongly OOF), as illustrated in panel (d). (a) Left: WSI of the lymph node from Figure 1b, exhibiting regional OOF artifacts. Right: predicted OOF classes. (b) Lymph node biopsy from colon cancer case (H&E stain). (c) Lymph node biopsy from breast cancer case (immunohistochemistry stain). (d) Color-coding of predicted OOF degrees and mapping to disk radii used in Bokeh blurring.

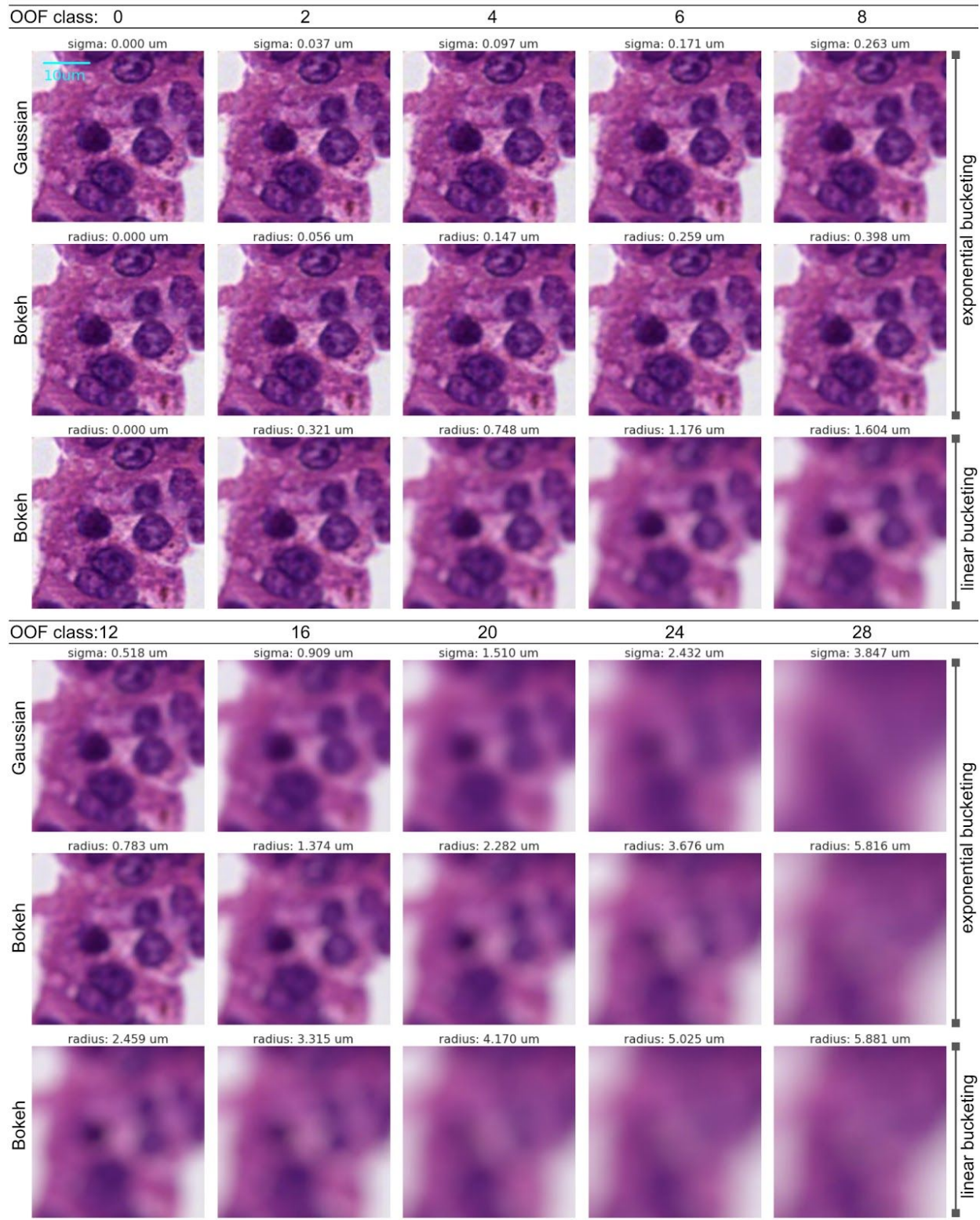

**Figure 4: Examples of different degrees of synthetic out-of-focus (OOF) used for training.** Gaussian or Bokeh blurring were applied at increasing levels to an in-focus image (see images at top left with σ=0 and radius=0). In order to maintain visually similar blur degrees for the same OOF class label, the σ values and Bokeh disk radii were aligned by minimizing the sum of squared differences between the blurred images.

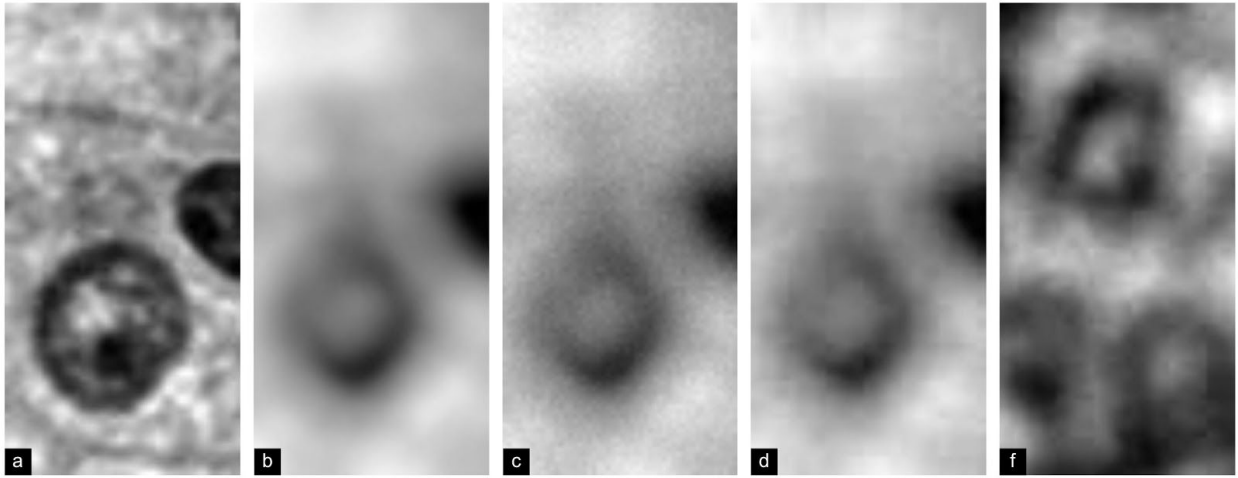

**Figure 5: Detailed comparison of real out-of-focus (OOF) to simulated OOF images**. (a) An in-focus image of a lymphocyte, shown in grayscale to improve visibility of artifacts: particularly grid-like artifacts from JPEG compression and Poisson noise from the sensor. (b) When synthetically generating OOF training examples, just applying Gaussian or Bokeh blurring alone smooths these artifacts. (c) Adding Poisson noise and (d) JPEG artifacts restores these artifacts, resulting in more realistic images. (e) An real OOF image of the same cell type for comparison.

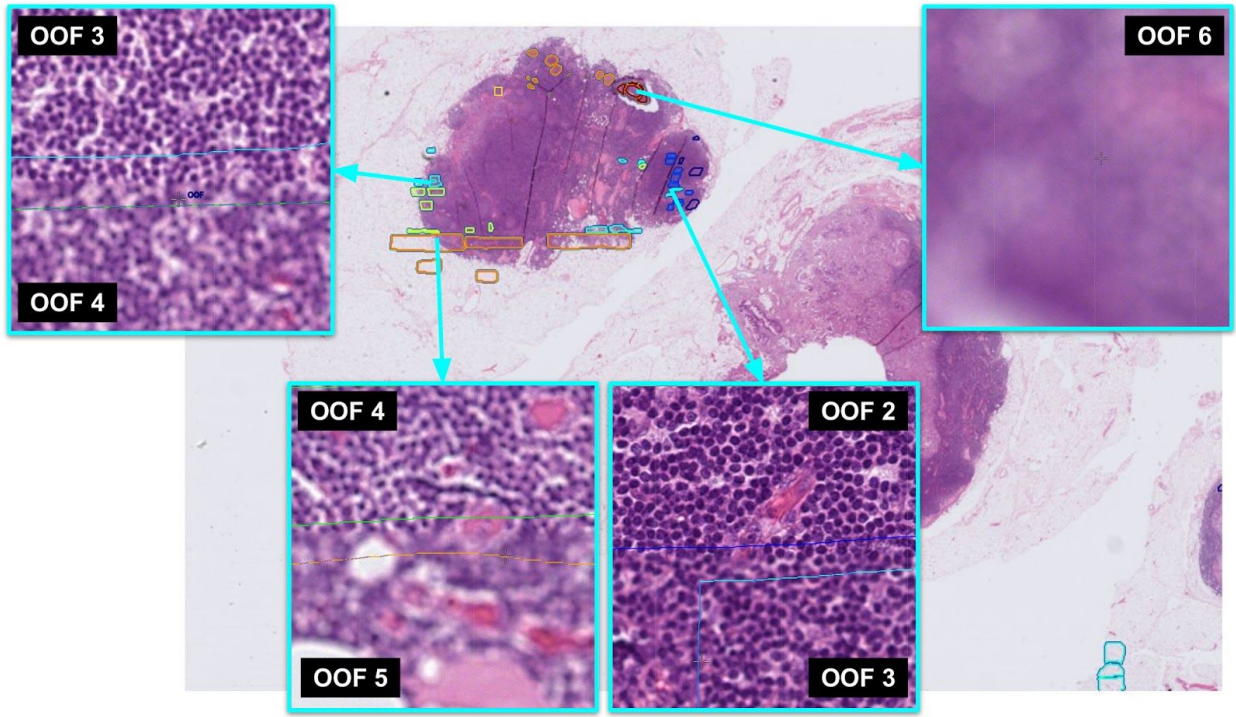

**Figure 6: Examples of out-of-focus (OOF) annotations by pathologist.** The pathologist identified, delineated and graded the regions highlighted in this lymph node from a colon cancer case (scanned at 40X on Leica AT2). Colors from the jet palette indicate the manual OOF annotation grade, ranging from 0 (in-focus, dark blue) to 6 (strong OOF, red).

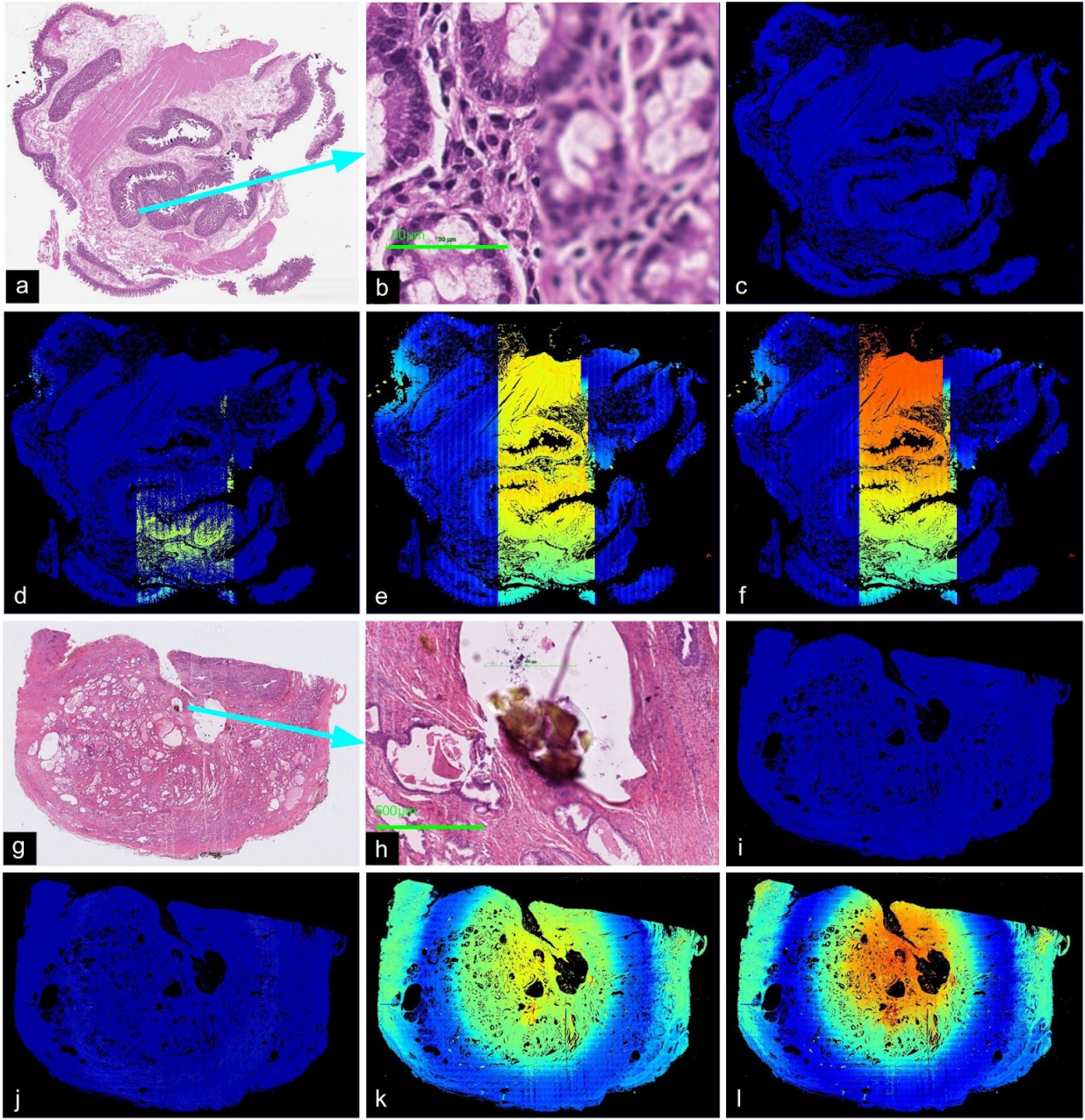

**Figure 7**: **Qualitative assessment of the impact of each step in our semi-synthetic OOF data generation process.** (a,g) Slide images of a duodenum and prostate specimen. (b,h) Magnified views of regions with OOF artifacts of panels a and g respectively. (c-f, i-l) Algorithm-predicted OOF classes; color map is shown in Figure 2(d). Four different configurations were applied (Table 2). (c,i) Algorithm trained with Gaussian blurring and brightness perturbations only. (d,j) Model trained with simulated JPEG artifacts in addition. (e,k) Algorithm trained with simulated Poisson noise in addition. (f,l): Algorithm trained with Bokeh blurring instead of Gaussian.

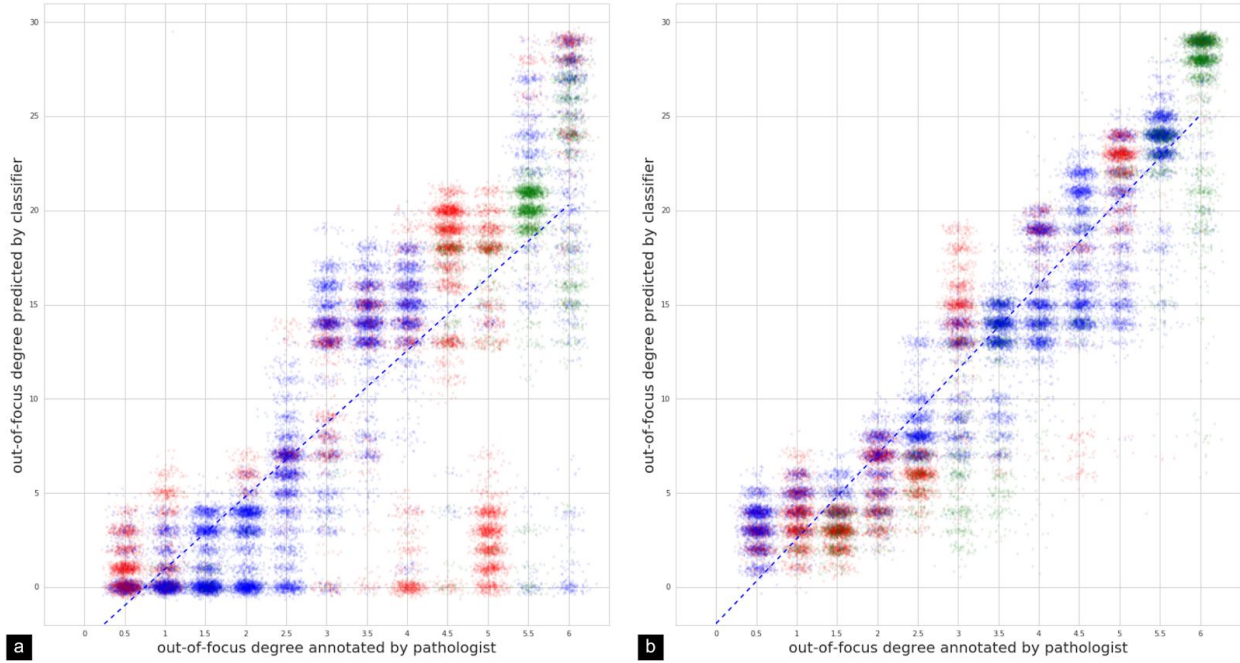

**Figure 8: Scatter plots of ConvFocus predictions against pathologist-annotated OOF grades.** Plot show 36,000 annotated image patches across 7 different slides scans per scanner model. To enable visualization of point density, small amounts of x,y jitter were added. Left: results for AT2 (Spearman's $\varrho$=0.808). Rights: results for S360 (Spearman's $\varrho$=0.936). Colors indicate patches from different specimens or stains; red: lymph node; blue: prostate; green: immunohistochemistry stained slides.

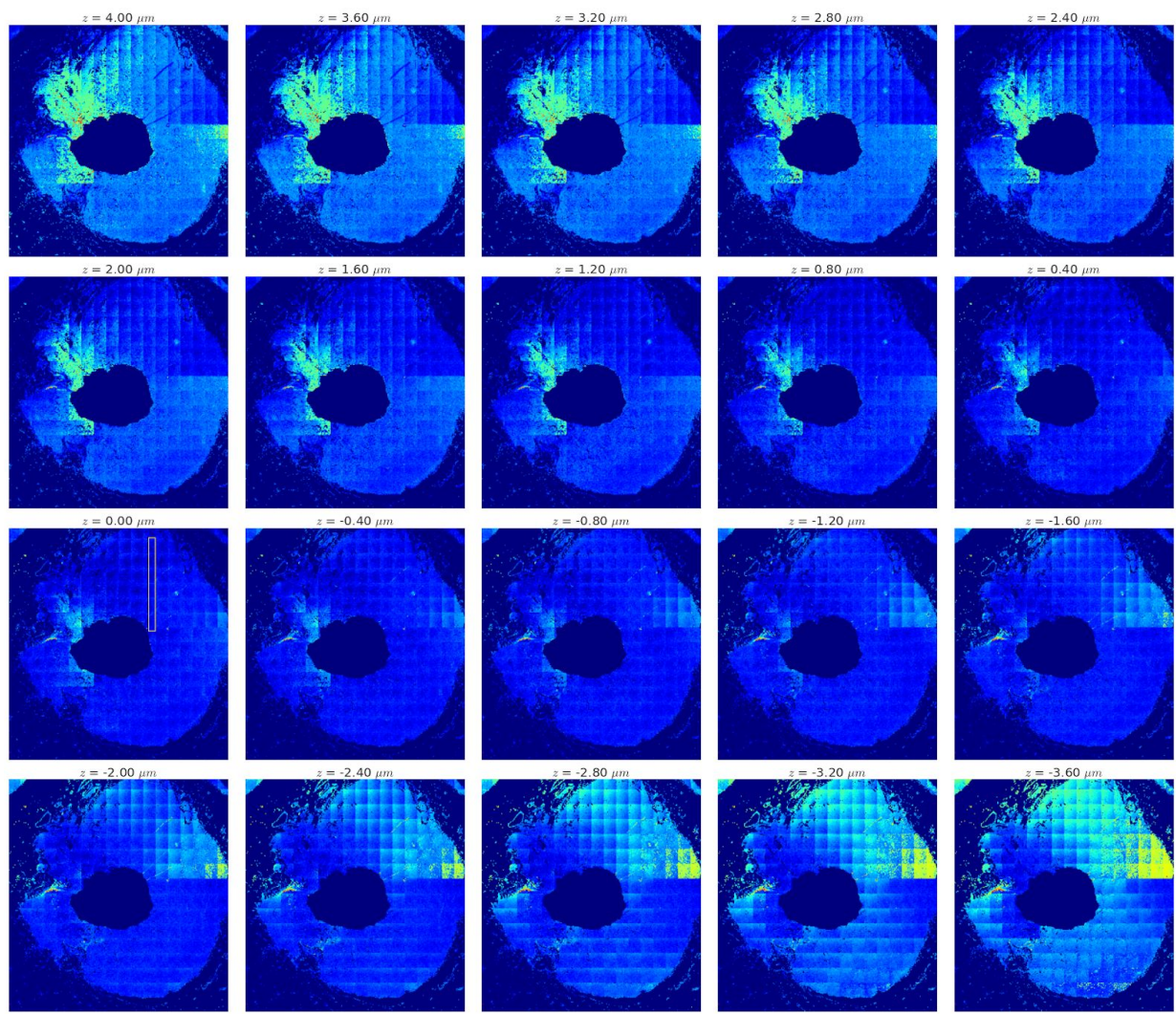

**Figure 9: ConvFocus predictions for z-stack scans from a lymph node biopsy of a colon cancer case.** This is the middle lymph node in Figure 3b. WSIs were acquired in z-stack mode on a NanoZoomer S360 ranging from +4μm (top left) to -3.6μm (bottom right) in 0.4μm increments. z=0μm indicates the scanner-determined "in-focus" plane. Plots of the OOF predictions for patches in the white rectangle in the z=0μm are shown in Figure 10.

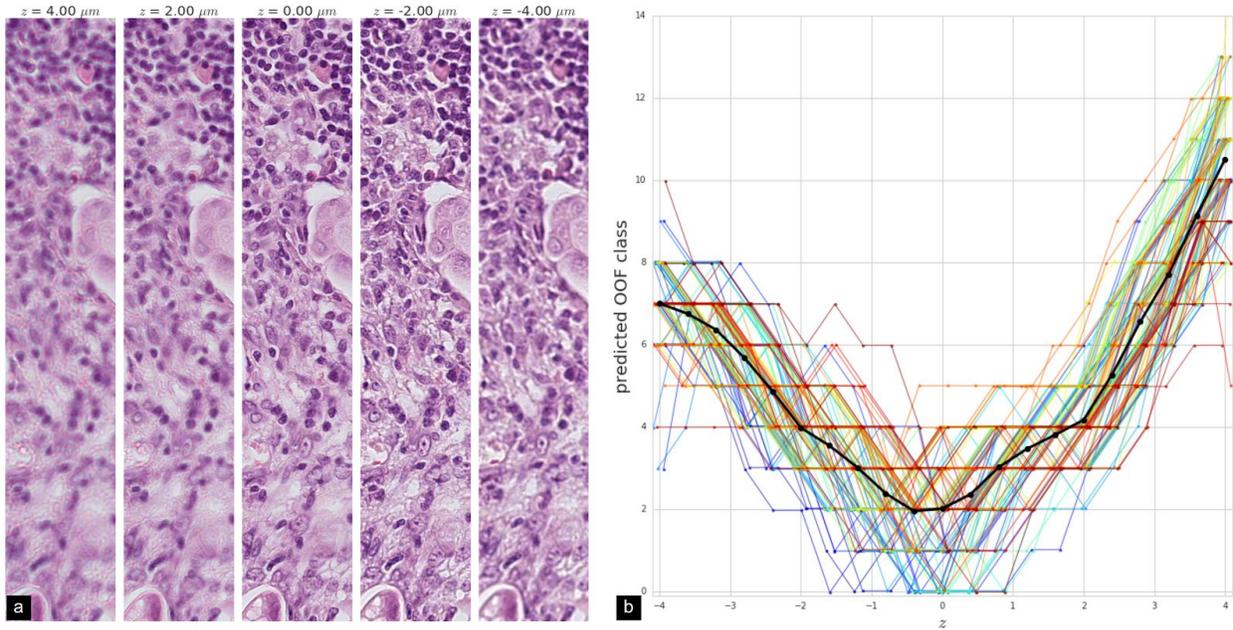

**Figure 10: Z-stack examples and OOF predictions along z-axis.** (a) Sample high magnification views at different z-levels. (b) Predicted OOF class plotted against z-level for 114 patches from the white rectangle in the z=0µm panel of Figure 9. Jitter has been added in the x and y dimensions for better visualization. Different colors are used to clearly indicate different patches. Mean class predictions across these patches are shown in black to highlight the average trend. Spearman's rank correlation rho between the absolute value of z and mean class predictions averages are: > 0.999 for $z \leq 0$ and 1.0 for $z \geq 0$ ($p < 0.001$ for both).

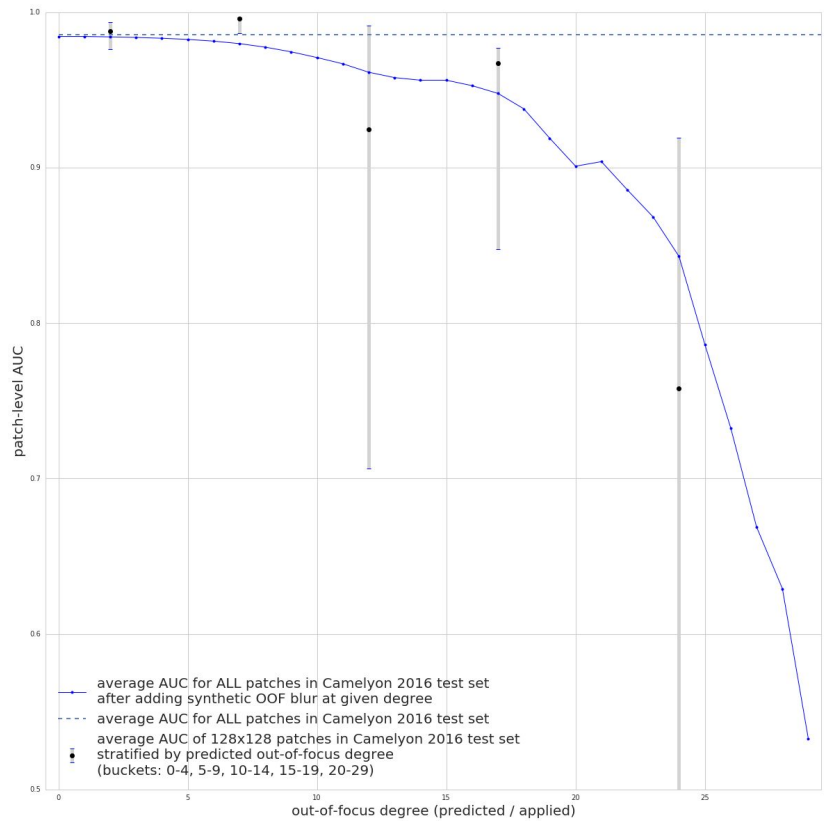

**Figure 11**: **Performance of a breast cancer metastasis detection algorithm as a function of OOF degree.** Patch-level area under receiver operating characteristic curve (AUC) was computed using annotations in the publically-available Camelyon 2016 challenge's test set. Performance degrades whether looking at real images of different OOF degrees (black dots with gray bars that indicate 95% confidence intervals), or synthetically blurred OOF images (blue line). Dotted line represents the patch-level AUC of the cancer detection algorithm across all patches (independent of OOF degree).

# Tables

| Surgical site | # slides | Surgical site | # slides |
|---|---|---|---|
| prostate, seminal vesicles | 3857 | submandibular gland | 51 |
| breast | 1622 | fallopian tube, ovary | 48 |
| uterus | 664 | ovary | 47 |
| thyroid | 524 | kidney | 23 |
| colon | 499 | uterus, fallopian tube, ovary | 22 |
| lymph node | 239 | products of conception | 21 |
| tonsil | 149 | appendix | 20 |
| vas deferens | 144 | jejunum | 12 |
| placenta | 93 | fetus | 10 |
| duodenum | 87 | liver | 8 |
| bladder | 87 | stomach, esophagus | 8 |
| stomach | 87 | aorta | 3 |
| omentum | 83 | testicle | 3 |
| ileum | 79 | spermatic cord | 2 |
| esophagus | 67 | vulva | 2 |
| breast, lymph node | 59 | gallbladder | 1 |
| colorectal | 58 | tonsil, uvula | 1 |
| skin | 54 | parotid | 1 |
| fallopian tube | 53 | | |

**Table 1: Surgical site information for a 8,135 slide subset of the training slides (for which this was available).**

| Confi-guration # | Description | Blur method | Spearman rank correlation | | Linear regression | | | |
|---|---|---|---|---|---|---|---|---|
| | | | | | slope | | intercept | |
| | | | AT2 | S360 | AT2 | S360 | AT2 | S360 |
| 1 | Default (only blur) | Gaussian | -0.274 | -0.134 | -0.023 | -0.211 | 2.504 | 1.949 |
| 2 | Configuration #1, plus simulated Poisson noise | Gaussian | -0.140 | -0.325 | -0.057 | -0.265 | 0.502 | 1.234 |
| 3 | Configuration #2, plus JPEG artifacts | Gaussian | 0.785 | 0.922 | 2.851 | 2.851 | 2.801 | 2.801 |
| **4** | **Configuration #3 (ConvFocus), replacing Gaussian with Bokeh blurring** | **Bokeh** | **0.808** | **0.936** | **3.903** | **4.408** | **-2.945** | **-1.730** |
| 5 | Configuration #4, replacing ConvFocus's simpler convolutional neural network with a more complex, modified Inception architecture | Bokeh | 0.773 | 0.934 | 3.889 | 4.504 | -3.024 | -1.969 |

**Table 2: Effects on performance of adding each step in our proposed semi-synthetic data generation process.** The final configuration illustrates that a more complex architecture is not required for this specific task.

# Appendix

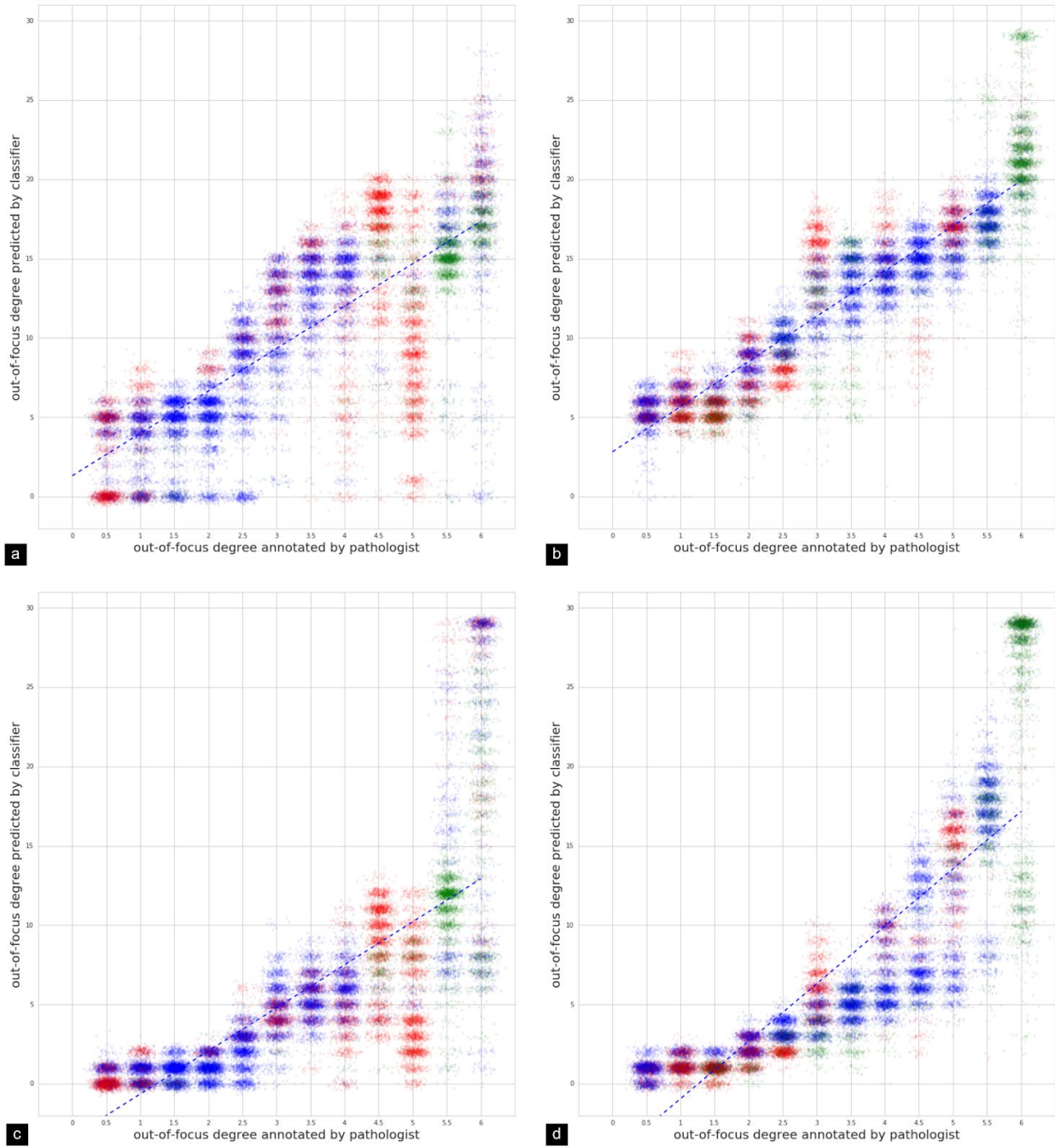

**Figure A1: Scatter plots of other configurations of OOF models against pathologist annotations** (compare with results in Figure 8). (a,b) Results of OOF model trained with Gaussian instead of Bokeh blurring (configuration #3 in Table 2). (c,d) Results of OOF model trained with Bokeh-blurring model but with a linearly increasing kernel size instead of exponential. Left column (a,c) indicates AT2 results; right column (b,d) indicates S360 results.

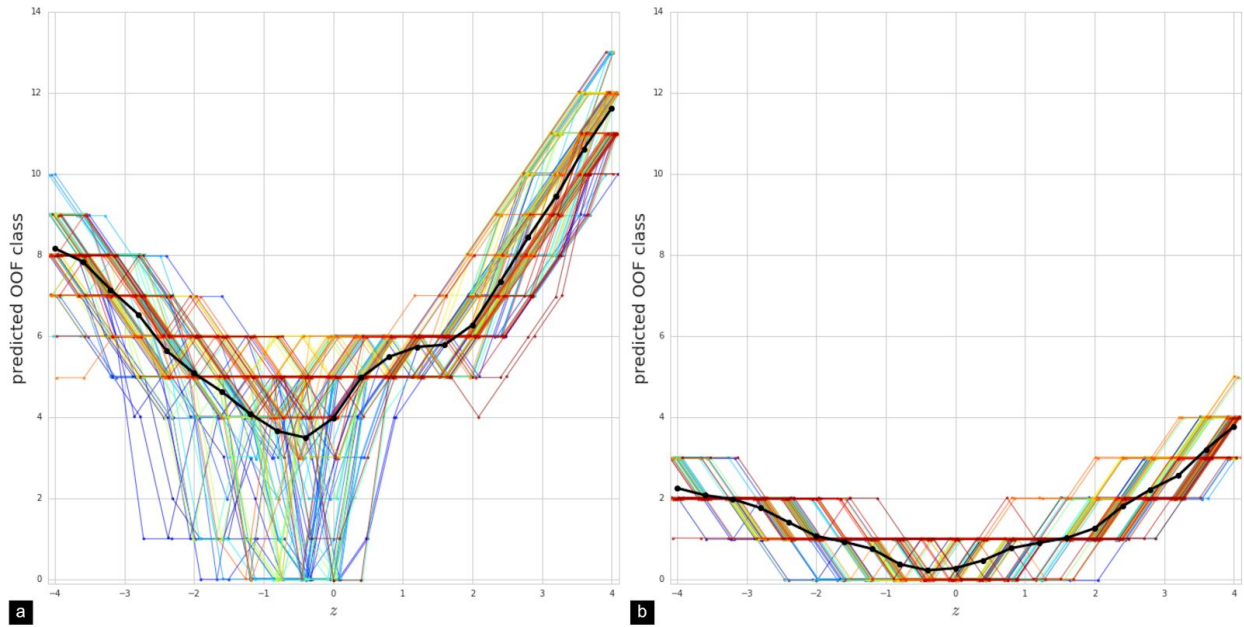

**Figure A2: Predicted OOF classes over z for patches sampled from the white rectangle in the z=0 panel of Figures A3, A4.** (a): Results of configuration trained with Gaussian instead of Bokeh blurring (configuration #3 in Table 2). (b): Results of configuration trained with Bokeh-blurring but with linearly increasing convolution mask size instead of exponential. Mean class predictions across these patches are shown in black to highlight the average trend.

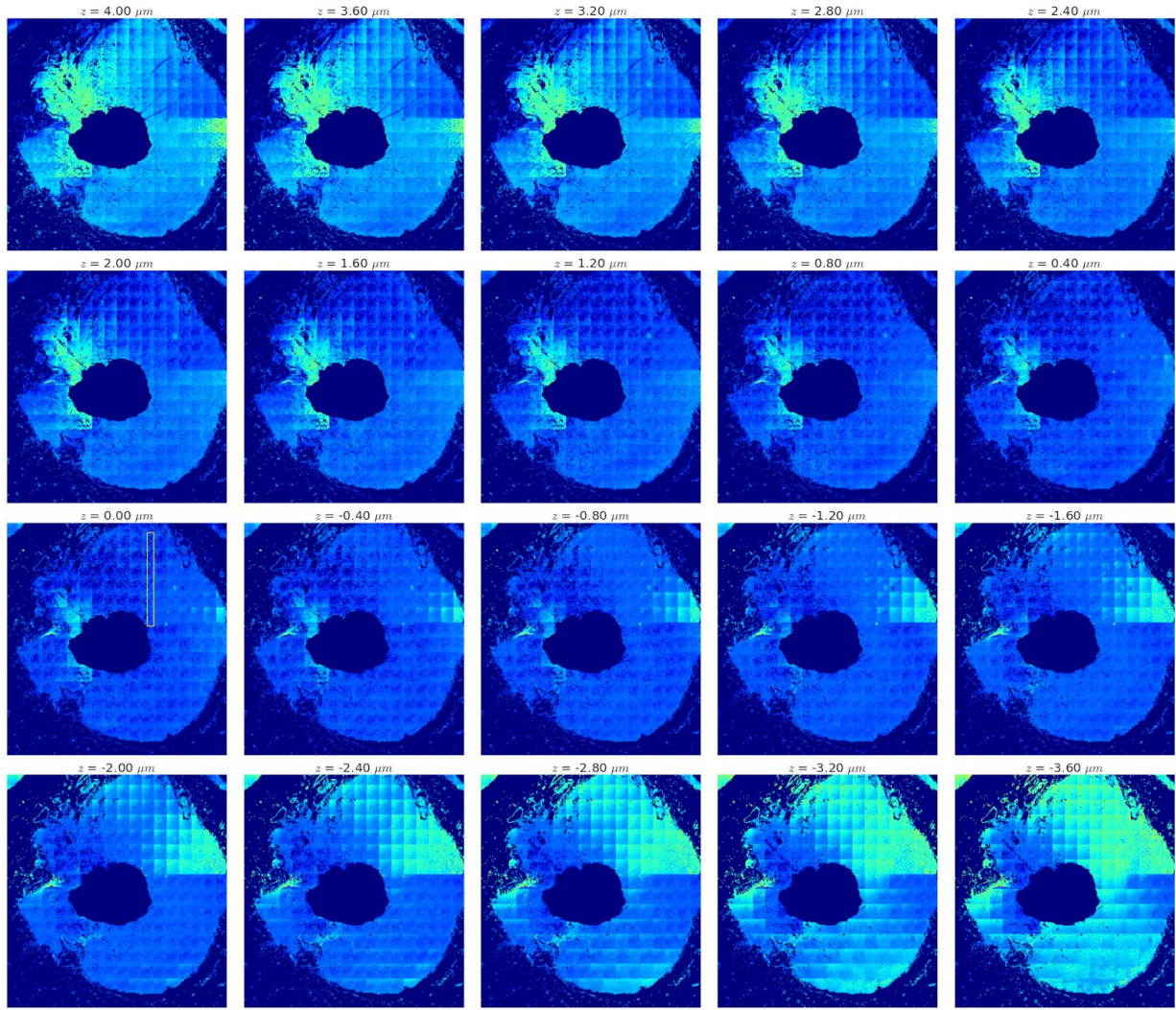

**Figure A3: Results of Gaussian-trained configuration on z-stack scans.** Corresponds to results shown in Figure A2a.

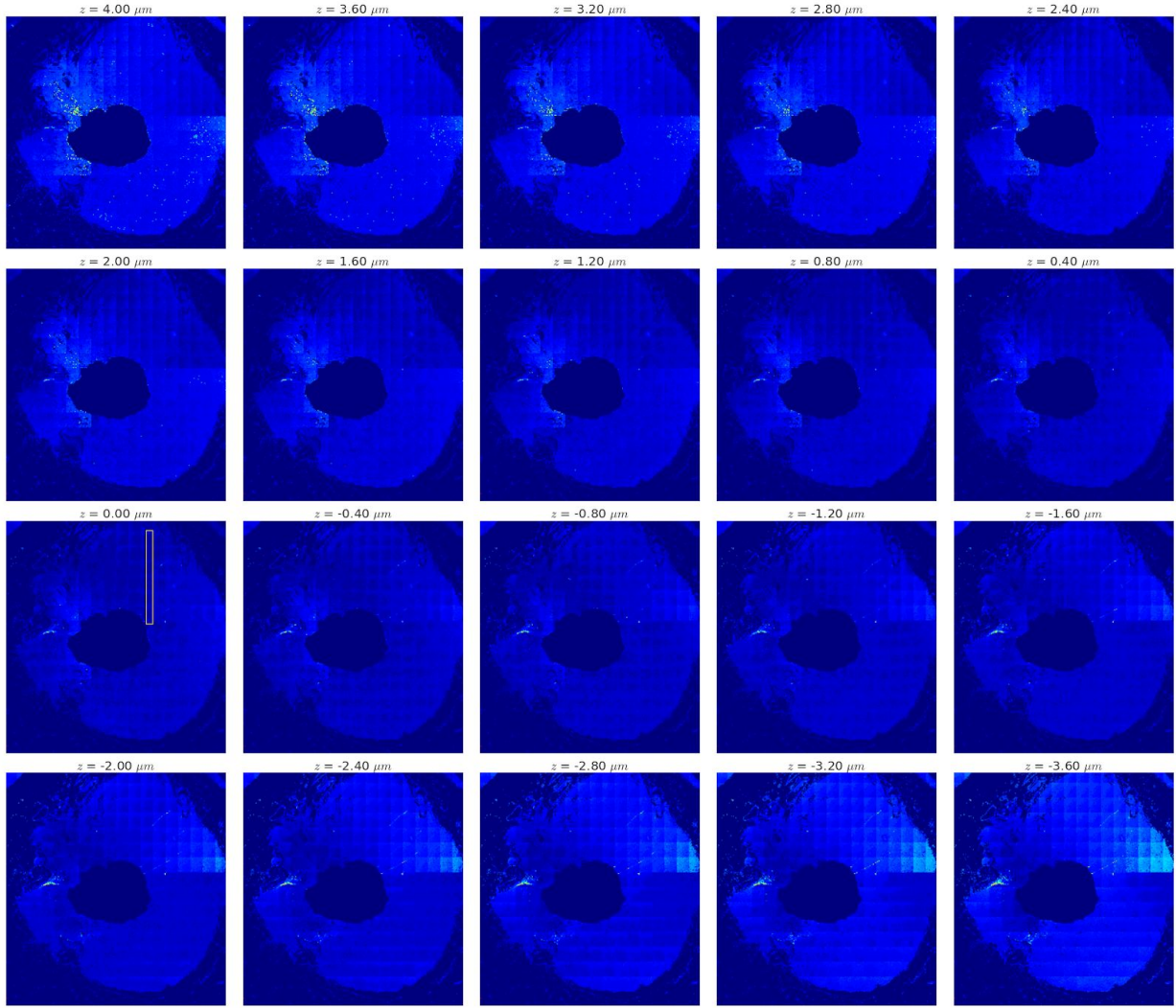

**Figure A4: Results of linear kernel size configuration on z-stack scans.** Corresponds to results shown in Figure A2b.